\pdfoutput=1
\documentclass[11pt]{article}

\usepackage[final]{acl}

\usepackage{times}
\usepackage{latexsym}

\usepackage[T1]{fontenc}
\usepackage[utf8]{inputenc}

\usepackage{microtype}

\usepackage{inconsolata}
\usepackage{multirow}
\usepackage{booktabs}
\usepackage{graphicx}
\usepackage[mathscr]{eucal}
\usepackage{amsmath}
\usepackage{amsfonts,amssymb}
\usepackage{tikz}
\usepackage{bbding}
\usepackage{color}
\usepackage{xcolor}
\usepackage{makecell}
\usepackage[misc]{ifsym}
\title{MCAD: Multi-teacher Cross-modal Alignment Distillation for efficient image-text retrieval}

\author{
Youbo Lei$^{1, *}$, Feifei He$^{2, *}$, Chen Chen$^{2, \textrm{\Letter}}$, Yingbin Mo$^{3}$,\\
{\bf Si Jia Li$^{2}$, Defeng Xie$^{2}$, Haonan Lu$^{2, \textrm{\Letter}}$} \\
$^1$Xi'an Jiaotong University \quad $^2$OPPO AI Center \quad $^3$University of Sydney
}

\begin{document}
\maketitle
\footnotetext[1]{*Equal contribution (Youbo.LEI@stu.xjtu.edu.cn)}
\footnotetext[2]{\textrm{\Letter} Corresponding authors (chenchen4@oppo.com; luhaonan@oppo.com).}

\begin{abstract}
Due to the success of large-scale visual-language pretraining (VLP) models and the widespread use of image-text retrieval in industry areas, it is now critically necessary to reduce the model size and streamline their mobile-device deployment. 
Single- and dual-stream model structures are commonly used in image-text retrieval with the goal of closing the semantic gap between textual and visual modalities. While single-stream models use deep feature fusion to achieve more accurate cross-model alignment, dual-stream models are better at offline indexing and fast inference.
We propose a Multi-teacher Cross-modality Alignment Distillation (MCAD) technique to integrate the advantages of single- and dual-stream models. By incorporating the fused single-stream features into the image and text features of the dual-stream model, we formulate new modified teacher similarity distributions and features. Then, we conduct both distribution and feature distillation to boost the capability of the student dual-stream model, achieving high retrieval performance without increasing inference complexity.
Extensive experiments demonstrate the remarkable performance and high efficiency of MCAD on image-text retrieval tasks. Furthermore, we implement a lightweight CLIP model on Snapdragon/Dimensity chips with only $\sim$100M running memory and $\sim$8.0ms search latency, achieving the mobile-device application of VLP models. 

\end{abstract}

\section{Introduction}

Image-text mutual retrieval is a fundamental problem of multimodal learning, whose primary objective is to bridge the semantic gap between visual and textual modalities, enabling accurate match of image (text) based on the given text (image). However, aligning and matching visual and textual information is non-trivial due to the differences in their representations and structures. 
%Images are represented as continuous visual features, while texts are represented as discrete language tokens.
In recent years, the rapid growth of large-scale paired vision-language datasets~\cite{schuhmann2021laion, schuhmann2022laion} has paved the way for the development of powerful models that can bridge the gap between visual and textual information. These models, known as vision-language pretraining (VLP) models, have shown remarkable capabilities in understanding both vision and language~\cite{radford2021learning, jia2021scaling, li2021align}.

Typically, the dual-stream architecture, \textit{e.g.}, CLIP~\cite{radford2021learning}, ALIGN~\cite{jia2021scaling}, facilitates autonomous processing of individual modalities through segregated streams, exhibiting inferior retrieval performance due to the lack of effective cross-modal feature fusion. Nevertheless, the disentanglement of image and text encoder enable fast retrieval speed. On the contrary, single-stream models integrate information from multiple modalities during encoding through a deep interaction module, \textit{e.g.}, transformer block~\cite{vaswani2017attention}, commonly leading to superior retrieval performance but sacrificing flexibility and resulting in extremely low retrieval speed. %Therefore, in industrial applications, dual-stream models are still the first choice. However, their significant size hinders practical deployment in lightweight scenarios, especially for mobile devices. 
Thus, despite their size hindering deployment in lightweight scenarios like mobile devices, dual-stream models remain the preferred choice in industrial applications.

In recent years, several works endeavor to transfer knowledge of large models into small models through distillation technology~\cite{fang2021compressing,wang2022efficientvlm,rao2023dynamic,ren2022leaner,wang2021distilled,miech2021thinking,lei2022loopitr,wu2023tinyclip,vasu2023mobileclip}. But they just consider soft-label, feature, or attention map distillation from one teacher or homogeneous teachers. The strategy of homogeneous, multi-teacher distillation has not yet been explored.
%As mentioned above, single-stream models fuse multi-modal features to achieve better visual-language understanding capability, but 
Among these works, a critical question is how to distill the knowledge of the single-stream models into efficient dual-stream models. Although DIDE~\cite{wang2021distilled} proposes to employ cross-modal attention distillation to transfer the knowledge of the ViLT~\cite{kim2021vilt} teacher to a CLIP student, this is not a universal method since other single-stream structures, \textit{e.g.}, ALBEF~\cite{li2021align} cannot discriminate explicit image and text features after cross-attention fusion so that the image-text attention maps are unavailable. Another work LoopITR~\cite{lei2022loopitr} considers only employ the single-stream output scores of top $k$ hard examples chosen by the dual-stream model to enhance the dual-stream model itself, which cannot excite the whole ability of the single-stream model. 
%{\color{cyan}On the other hand, recent studies on multi-teacher frameworks predominantly explored single-modal domains such as voice, text, and images, utilizing teacher models of identical structures for effective knowledge distillation to student models through shared inputs and combined intermediate and final teacher outputs\cite{pham2023collaborative,huang2023ensemble}.}
So the integration of single-stream and dual-stream teachers is a non-trivial challenge.
In this paper, we are motivated to propose a Multi-teacher Cross-modal Alignment Distillation (MCAD) method to make full use of the information fusion ability of the single-stream model and the large-scale parallel training advantage of the dual-stream model. Specifically, after extracting features through the frozen single- and dual-stream teacher models, we apply different learnable projection layers to align image or text features from different latent spaces, as shown in Fig.~\ref{fig:architecture_cs}. Finally, we employ similarity distribution and feature distillation based on the newly-formulated fused features to boost the performance of the dual-stream student model, as shown in Fig.~\ref{fig:model_overview}. In summary, our main contributions are as follows:

\begin{itemize}
\setlength{\leftmargin}{-1pt}
\setlength{\topsep}{0pt}
\setlength{\itemsep}{0pt}
\setlength{\parsep}{0pt}
\setlength{\parskip}{0pt}
    \item We propose a single- and dual-stream multi-teacher distillation algorithm to enhance the cross-modal retrieval ability of a light-weight CLIP-like dual-stream model.
    \item Comprehensive experiments on different datasets and networks demonstrate that our method is a model-agnostic general framework that can achieve superior performance both in zero-shot and fine-tuning settings.
    \item By using MobileViTv2~\cite{mehta2023separable} and TinyBERT~\cite{tinybert2020} as the image and text encoder, respectively, we compress a 400M large CLIP model onto Snapdragon/Dimensity chips, achieving merely 25.9M model size, $\sim$100M running memory, and $\sim$8.0ms retrieval latency.
\end{itemize}

\section{Related Work}
\subsection{Image-Text Retrieval with VLP}
Image-text retrieval (ITR) has attracted increasing attention in recent years.
%The early paradigm for ITR involves fine-tuning pre-trained models specialized in computer vision and natural language processing domains~\cite{8100043}, respectively. In recent years,
%With the widespread adoption of attention mechanism~\cite{vaswani2017attention} in computer vision~\cite{liu2021swin, dosovitskiy2021an},
In recent years, cross-modal pre-training has been extensively studied and applied to ITR~\cite{ITM2019, vilbert2019, chen2020uniter, wang2022simvlm}. The model structure can be roughly classified into two categories: single-stream and dual-stream. Single-stream models jointly encode images and text through a deep interaction module and output a fused feature. Early algorithms~\cite{vilbert2019} employ object detectors~\cite{girshick2015fast, ren2015faster} to extract image features, which usually ignore important background information. Then, ViLT~\cite{kim2021vilt,diao2021similarity} unifies image and text extractor as Transformer~\cite{vaswani2017attention} to make full use of all information. The models, however, depend on a cross-modal Transformer encoder to fuse visual and textual signals at the same time across layers, which necessitates a large compute budget and slows down inference speed. Even though some trade-off approaches, \textit{e.g.}, ALBEF~\cite{li2021align}, employ separate image and text encoders prior to hard example fusion, their top $k$ re-ranking strategy is still far from being implemented in real time.

On the contrary, the dual-stream model mainly focuses on learning how to align visual and textual features obtained from independent encoders. Since only a light-weight interaction module (usually a MLP or dot product) is applied to image and text features, dual-stream structure allows for contrastive learning on billions of examples, including CLIP~\cite{radford2021learning} and ALIGN~\cite{jia2021scaling}. Thanks to the shadow interaction module, all visual or textual features can be pre-calculated and stored offline, leading to a fast retrieval speed. Nevertheless, due to a lack of deep cross-model fusion, the visual-language understanding ability of dual-stream models is inferior to that of single-stream models, resulting in lower retrieval accuracy. Hence, we are inspired to transfer the advantages of single- and dual-stream models into a compressed, lightweight model through our proposed innovative distillation technique.

\subsection{Knowledge Distillation for VLP}
Knowledge Distillation~\citep{hinton2015distilling} is a method of transferring knowledge from a teacher model to a student model, which can effectively improve the performance of the student model \citep{Lan2020ALBERT, DBLP2019,touvron2021training,9711361}.
In the multimodal distillation area, a group of approaches considers transferring knowledge from large models into small models with the same architecture, either both single-stream models~\cite{fang2021compressing,wang2022efficientvlm,rao2023dynamic} or dual-stream models~\cite{ren2022leaner, wu2023tinyclip} by using logit, feature, or attention map distillation. DistillVLM and EfficientVLM~\cite{wang2022efficientvlm} propose attention map distillation and hidden feature distillation for object-detection-based and ALBEF-like~\cite{li2021align} single-stream VLP model compression, respectively. TinyCLIP~\cite{wu2023tinyclip} trains lightweight CLIP models via cross-modal affinity mimicking (similarity distribution distillation) and weight inheritance.

Another group of methods employ single-stream models to improve performance of dual-stream models~\cite{wang2021distilled,miech2021thinking,lei2022loopitr}. DIDE~\cite{wang2021distilled} applies cross-model attention distillation to transfer knowledge of a single-stream ViLT~\cite{kim2021vilt} teacher model into a CLIP-like dual-stream student model. LoopITR~\cite{lei2022loopitr} proposes a mutual-loop enhancement strategy to distill dual-stream models by top hard samples of single-stream models. Thinking fast and slow~\cite{miech2021thinking} improves dual-stream model performance by single-stream model via logit distillation.

As multi-teacher distillation~\citep{yang2020model, gou2021knowledge, zhao2022enhanced, zhang2023autodisc} has been generally regarded as an effective approach to improving student models, MobileCLIP~\citep{vasu2023mobileclip} proposes to employ ensemble of $K$ CLIP models as a strong teacher. To the best of our knowledge, our MCAD is the first work that uses heterogeneous multi-teachers to distill the advantages of single- and dual-stream models into a lightweight student VLP model.
%the aforementioned methods only utilize a single teacher model without considering multi-teacher distillation, especially for models with distinct structures. \citep{nie2023lightclip}

\section{Method}

\begin{figure}[htb]
	\centering
\setlength{\abovecaptionskip}{0.2em}
\setlength{\belowcaptionskip}{0.2em}
    \includegraphics[width=0.45\textwidth]{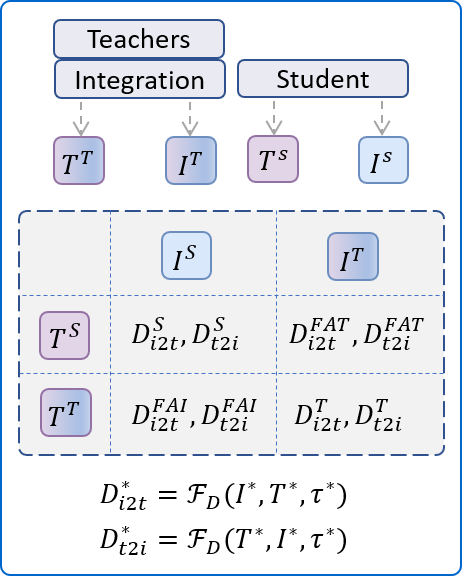}
    \caption{An overview of our MCAD framework. $(I^T, T^T)$ and $(I^S, T^S)$ represent the (image, text) feature pair output by teachers and the student, respectively. $D^*_{i2t}$ represents the similarity distribution of image-to-text, while $D^*_{t2i}$ denotes that of text-to-image. $D^S_*$ indicates the distribution matrix produced by the student, while $D^T_{*}$ depicts that derived from the aggregated teachers. Additionally, $D^{FAI}_{*}$ denotes the softmax output after cross-feature alignment between the student's image feature and the teachers' text feature, while $D^{FAT}_{*}$ represents the corresponding operation after aligning the student's text feature to the teachers' image feature.
    }
\label{fig:model_overview}
\end{figure}

\subsection{Preliminary}
We first define the general form for calculating the similarity distribution matrix and the KL divergence loss, then we will introduce the distribution matrices shown in Fig.~\ref{fig:model_overview}.

The general form to calculate the image-text similarity distribution matrix can be denoted as:
\begin{equation}
\mathcal{F}_{D}(I, T, \tau) = \operatorname{softmax}[ (I T^{\top})/ \tau],
\label{equ:infonce}
\end{equation}
where $\operatorname{softmax}(\cdot)$ represents the softmax function that operates in the last dimension, $I, T$ denote the normalized image and text representations, respectively, with shape $[n, d]$, where $n$ is the batch size and $d$ is the output dimension, and $\tau$ is a temperature parameter.
Moreover, the row-wise KL divergence between two distribution matrices $D$ and $\hat{D}$ can be denoted as:
 \begin{equation}
    \mathcal{F}_{KL}(D, \hat{D}) = \sum _l \operatorname{KL}(D_l || \hat{D}_l ),
    \label{equ:kl}
\end{equation}
where $l$ indicates the row index.

Fig.~\ref{fig:model_overview} shows an overview of the MCAD framework, which combines single- and dual-stream models at the token level. Given $n$ image-text pair inputs in a batch, $\{(i_j, t_j)\}_1^n$, we will get image representations $I^T\in \mathbb{R}^{n\times d}$ and text representations $T^T \in \mathbb{R}^{n\times d}$ after feeding the output of multiple teachers to the integration module, which will be detailedly discussed in Sec.~\ref{sec: multiteacher}. Moreover, the student's image encoder and text encoder output the image representation $I^S \in \mathbb{R}^{n\times d}$ and text representation $T^S \in \mathbb{R}^{n\times d}$, respectively.
After that, several distribution matrices shown in Fig.~\ref{fig:model_overview} can be expressed as:
\begin{equation}
\begin{split}
    & D^{S}_{i2t} = \mathcal{F}_{D}(I^S, T^S, \tau^S), \\
    & D^{T}_{i2t} = \mathcal{F}_{D}(I^T, T^T, \tau^T), \\
    & D^{FAI}_{i2t} = \mathcal{F}_{D}(I^S, T^T, (\tau^T+\tau^S)/2), \\
    & D^{FAT}_{i2t} = \mathcal{F}_{D}(I^T, T^S, (\tau^T+\tau^S)/2), \\
\end{split}
\end{equation}
where $D^S_{i2t}$ denotes the similarity distribution of image-to-text output by student, while that of text-to-image simply involves swapping the input positions, \textit{i.e.}, $D^S_{t2i} = \mathcal{F}_{D}(T^S, I^S, \tau^S)$. The detailed description of $D^*_*$ can be found in the caption of Fig.~\ref{fig:model_overview}.
%It's worth noting that the output distributions of the integrated teachers also need to be aligned 

\subsection{Multi-teacher Cross-modal Alignment Distillation}
\label{sec: MCAD}

\textbf{Dual-stream target.} Assuming a collection of $n$ image-text pairs $\{(i_j, t_j)\}_1^n$ in a batch, the text and image features of the dual-stream teacher model are denoted as $I^{DS} \in \mathbb{R}^{n\times d}$ and $T^{DS} \in \mathbb{R}^{n\times d}$, respectively, and the target similarity distribution matrix can be expressed as:
\begin{equation}
\begin{split}
    & D^{DS}_{i2t} = \mathcal{F}_{D}(I^{DS}, T^{DS}, \tau^{DS}),  \\
    & D^{DS}_{t2i} = \mathcal{F}_{D}(T^{DS}, I^{DS}, \tau^{DS}), 
\end{split}
\label{equ:dual_target}
\end{equation}
where $\tau^{DS}$ denotes the temperature of the dual-stream model. 

\textbf{Single-stream target.} Besides the straightforward format of dual-stream target distributions, we also need to calculate the single-stream target. Subsequently, the indices of the top $k$ similarity scores are first computed based on Eq.~(\ref{equ:dual_target}), which can be represented as:
\begin{equation}
\begin{split}
    & P_{i2t} = \operatorname{topK\_indices}(D_{i2t}^{DS}), \\
    & P_{t2i} = \operatorname{topK\_indices}(D_{t2i}^{DS}),
\end{split}
\label{equ:dual_topk}
\end{equation}
where $P_{i2t}$ denotes the indices of each image and the top $k$ texts that are similar to it, while $P_{t2i}$ represents the indices of each text and its $k$ most similar images.
Then, we recalculate the scores of the top $k$ image-text pairs by the single-stream model, \textit{e.g.,} ALBEF~\cite{li2021align}. We assume that the score matrices output by the single-stream model are $D^{SS}_{i2t} \in \mathbb{R}^{n\times n}, D^{SS}_{t2i} \in \mathbb{R}^{n\times n}$, which are calculated as:
\begin{equation}
\begin{split}
    (D^{SS}_{i2t})_{l,m} = f_{SS}(i_l, t_m), \quad(l,m) \in P_{i2t}, \\
    (D^{SS}_{t2i})_{l,m} = f_{SS}(i_m, t_l), \quad(l,m) \in P_{t2i}. 
\end{split}
\label{equ:single_stream_dist}
\end{equation}
In general, a single-stream model will usually output a similarity score for the current image-text pair. It should be noted that in matrix $D^{SS}_{i2t}$ and $D^{SS}_{t2i}$, only $D^{SS}_{i2t}[P_{i2t}] \in \mathbb{R}^{n\times k}, D^{SS}_{t2i}[P_{t2i}]\in \mathbb{R}^{n\times k}$ are computed, and we only care about this part.

\textbf{Loss function.} The objective of this paper is to introduce the MCAD technique for effectively merging single- and dual-stream models. The ultimate goal is to enable effective knowledge transfer from multiple teachers to the student network. We adopt a dual-stream architecture for the student network, which results in improved retrieval speed for image-text tasks. In doing so, the proposed method can be more conveniently deployed on mobile devices. In this study, the uniform loss function is denoted as:
\begin{equation}
\label{eq:total_loss}
    \mathcal{L}_{total} = \mathcal{L_{TDD}} + \mathcal{L_{TFD}},
\end{equation}
where $\mathcal{L_{TDD}}$ denotes the loss function of target distribution distillation 
(TDD), and $\mathcal{L_{TFD}}$ denotes the target feature distillation (TFD).

First, the $\mathcal{L_{TDD}}$ of multi-teachers can be expressed as:
\begin{equation}
    \begin{split}
    \mathcal{L_{TDD}}: \mathcal{L}_{MT} &= f_{MT}(D^S_{i2t}, D^S_{t2i})
        \\&+  f_{MT}(D^{T}_{i2t}, D^{T}_{t2i}),
    \end{split}
    \label{equ:multi_teacher_tdd}
\end{equation}
where $f_{MT}$ is a loss function that measures the KL divergence between the output and the target distribution, including dual-stream and single-stream targets as mentioned before. Importantly, the second term brings the output similarity distribution of the integration module (discussed in Sec.~\ref{sec: multiteacher}) close to the target distribution, which can be viewed as a regularization of the integration module. 
Second, the $\mathcal{L_{TFD}}$ of multi-teachers is denoted as:
\begin{equation}
    \begin{split}
        \mathcal{L_{TFD}}: \mathcal{L}_{MT\_FA} &= f_{MT}(D^{FAI}_{i2t}, D^{FAI}_{t2i})  \\
    &+ f_{MT}(D^{FAT}_{i2t}, D^{FAT}_{t2i}), 
    \end{split}
    \label{equ:multi_teacher_tfd}
\end{equation}
where the two terms bring the representation of the student output close to the fused feature. 
%Notice that the fused feature is still regularized by the dual- and single-stream target distributions.

% The optimal target distribution is derived through the combination of single- and dual-stream model outputs.  Given n image text pairs , assume that the text representation and image representation of the output of dual-steam are, the distribution of the similarity matrix are denoted $D^{DS}_{i2t}, D^{DS}_{t2i}$. After that, we calculate top $k$ according to the similarity matrix, which can be denoted as follow:

Finally, the core loss function $f_{MT}$ is defined as:
\begin{equation}
\begin{split}
     f_{MT}(D^*_{i2t}, D^*_{t2i}) =
     \mathcal{F}_{KL}(D^*_{i2t}, D^{DS}_{i2t})& \\
    +\mathcal{F}_{KL}((D^*_{t2i}, D^{DS}_{t2i})& \\
    +\mathcal{F}_{KL}(\sigma(D^*_{i2t}[P_{i2t}]), \sigma(D^{SS}_{i2t}[P_{i2t}]))&\\
    +\mathcal{F}_{KL}(\sigma(D^*_{t2i}[P_{t2i}]), \sigma(D^{SS}_{t2i}[P_{t2i}]))&,\\
\end{split}
\label{equ:core_loss}
\end{equation}
where $* \!\in\! \{S, T, FAI, FAT\}$ and $\sigma(\cdot)$ is a normalization method. When given a matrix $D^* \in \mathbb{R}^{n\times k}$, the normalization method can be expressed as:
\begin{equation}
    \begin{split}
        \sigma(D^*_{l,m}) =   \frac{D^*_{l,m}}{\sum_{v=1}^{k} D^*_{l,v}}, \\ l \in [1,..,n], m \in [1, ..., k]
    \end{split}
\label{eq:L1norm}
\end{equation}
Here, we don't directly align the student output with the integrated-teacher output. Instead, we align the output of the student and the integrated teacher with the dual- and single-stream teacher simultaneously, as Eq.~(\ref{equ:multi_teacher_tdd})--(\ref{equ:core_loss}) show. Since the integrated teacher also contains learnable parameters (will be introduced in Sec.~\ref{sec: multiteacher}), we regard the dual- and single-stream as pivots to align the student and the integrated teacher. 
% 这里要解释一下为啥这么设计。实际上不是用student的output对齐integrated-teacher的output，而是同时对齐dual-stream和single-stream的output。因为integrated-teacher有g1-g4的训练参数，没约束的话输出是乱的，不能用来对齐。这里把integrated-teacher的output也和dual-stream和single-stream同时做了对齐，所以相当于把dual-stream和single-stream做了桥梁，让integrated-teacher和student对齐。

% \begin{equation}
%     \begin{split}
%     (D^{SS}_{i2t})_{l,m} = \frac{(O_{i2t})_{l,m}}{\sum_v^{(l,v) \in P_{i2t\_indices}} (O_{i2t})_{l,v}}, \\(l,m) \in P_{i2t\_indices} \\
%     (D^{SS}_{t2i})_{l,m} = \frac{(O_{t2i})_{l,m}}{\sum_v^{(l,v) \in P_{t2i\_indices}} (O_{t2i})_{l,v}}, \\
%     (l,m) \in P_{t2i\_indices} \\
% \end{split}
% \label{equ:single_stream_dist}
% \end{equation}
% where $D^{SS}_*$ represent the distribution matrix calculated by ALBEF. 

% After that, we introduce Cross Entropy (CE) to calculate the difference between the student output and the target. The Cross Entropy loss between two vector can be denoted as follow:
% \begin{equation}
%     \begin{split}
%         CE((D^*_*)_l, (D^{C}_*)_l) = - \sum_{m=1}^N (D^{C}_*)_{l,m} log ((D^*_*)_{l,m}) 
%     \end{split}
% \end{equation}

% However, due to the limitation of computing resources, we cannot calculate all the image and text pairs. To calculate the loss between single-stream and the student's output, the CE loss function needs to be modified as follows:
% \begin{equation}
% \begin{split}
%         & CE_{SS}((D^*_*)_l, (D^{SS}_*)_l) =   -\sum_m^{(l,m) \in P_{*\_indices}} \\ & \left[(D^{SS}_*)_{l,m}   log \frac{(D^*_*)_{l,m}}{\sum_{v}^{(l,v) \in P_{*\_indices}} (D^*_*)_{l,v}} \right]
% \end{split}
% \end{equation}

\subsection{Multi-teacher Integration}
\label{sec: multiteacher}

\begin{figure}[t]
\setlength{\abovecaptionskip}{0.2em}
\setlength{\belowcaptionskip}{0.2em}
	\centering
    \includegraphics[width=0.45\textwidth]{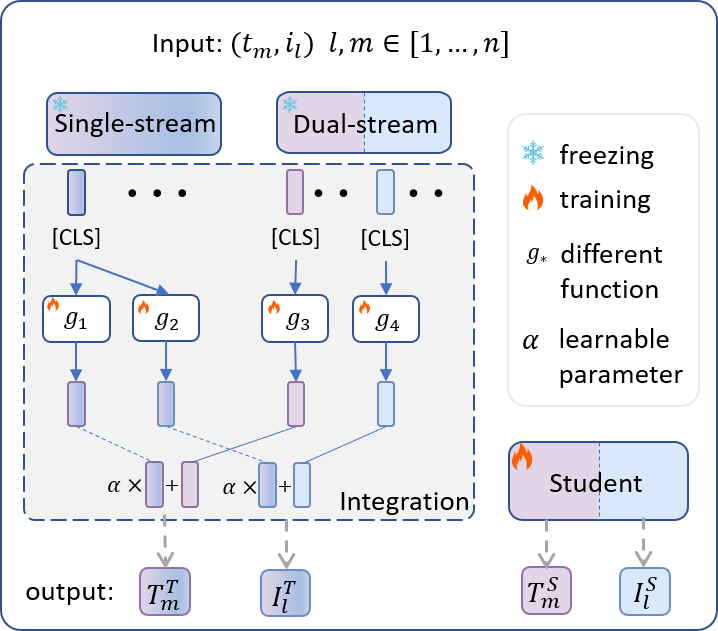}
    \caption{Details of the integration module}
	\label{fig:architecture_cs}
\end{figure}

To better utilize the features of multi-models, we propose a framework to integrate the output of different models, which is shown in Fig.~\ref{fig:architecture_cs}. When giving an image-text pair $(i_l, t_m), l, m \in [1,..., n]$, suppose that $I^{DS}_l, T^{DS}_m$ represent the ``CLS'' token output by dual-stream's image encoder and text encoder, respectively, and $H^{SS}_{m-l}$ represents the ``CLS'' token output by the single-stream model.
The ``CLS'' token outputs by multiple teachers are first projected into other vectors by a different function, $g_*$. Especially, although the single-stream model has only one ``CLS'' token, it still has to be projected to different spaces using two different functions, \textit{i.e.,} $g_1, g_2$. And in this study, the $g_1, g_2$ function can be denoted as follows:
\begin{equation}
    g_{1/2}(\cdot) \!=\! \begin{cases}
    f_{P}(H^{SS}_{m-l}), \!&\! if \ (l,m) \!\in\! P_{i2t} \\ \!&\! or\ (m,l) \!\in\! P_{t2i}\\
    0, \!&\! if \ (l,m) \!\notin\! P_{i2t} \\ \!&\! and \ (m,l) \!\notin\! P_{t2i},\\
    \end{cases}
\label{eq:projection_layer}
\end{equation}
where $f_{P}$ represents a projection layer. Moreover, $g_1$ and $g_2$ play the role of a gate. Because we fuse single- and dual-stream models to adjust the distribution of the top $k$, all we need to do is to fuse two teachers' features only on the top $k$.

Finally, the output of the text representation $T^T_m$ and image representation $I^T_l$ can be expressed as:
\begin{equation}
\begin{split}
    T^T_m = \operatorname{norm}(g_3(T^{DS-T}_m) + \alpha \cdot g_1(H^{SS}_{m-l})) \\
    I^T_l = \operatorname{norm}(g_4(I^{DS-I}_l) + \alpha \cdot g_2(H^{SS}_{m-l})),
\end{split}
\end{equation}
where $\alpha$ is a learnable parameter, and $\operatorname{norm}$ represents the $\ell_2$ normalization operator.

\section{Experiments}
\subsection{Datasets}
% \commentYb{It is reasonable for us to call it fine-tune?}
We utilize existing image-text pair datasets to verify our method, including MSCOCO~\citep{lin2014microsoft}, Conceptual Captions (CC)~\citep{sharma2018conceptual}, SBU captions~\citep{ordonez2011im2text}, and Flickr30K~\citep{7410660}. To test the zero-shot capability of our method, we only combine CC and SUB as training datasets, while in fine-tuning experiments, we use all four data datasets during training. For validation and testing, we utilize the standard split~\citep{7298932} of COCO and Flickr. More details of the datasets and training hyper-parameters are presented in Appendix~\ref{app:dataset_statistics} and \ref{app:training_details}, respectively.

%\subsection{Zero-shot Experiments}
\subsection{Baselines and Components}
\label{exp:baseline_component}
In this paper, as shown in Eq.~(\ref{eq:total_loss}), we propose a general loss function by dividing it into two parts: target distribution distillation ($\mathcal{L_{TDD}}$) and target feature distillation ($\mathcal{L_{TFD}}$) losses. We consider the first component to be the process of allowing the student's image-text similarity output to approximate a desired distribution. In terms of the second component, we can align the student's feature with the teacher's feature by following different constraints. Several prior works can be viewed as special cases of the general form proposed in Eq.~(\ref{eq:total_loss}). For the target distribution distillation, the categories can be summarized as follows:

\textbf{Ground truth.} Given $n$ image-text pairs $\{(i_l, t_l)\}_1^n$, the student model outputs two matrix $D^S_{i2t} \in \mathbb{R}^{n \times n}, D^S_{t2i} \in \mathbb{R}^{n \times n}$, and the ground truth can be denoted as $D^{GT}$, which is an identity matrix. Then, the loss function using the ground truth as the target distribution can be expressed as:
\begin{equation}
    \begin{split}
        \mathcal{L}_{TDD}: \mathcal{L}_{GT} &= \mathcal{F}_{KL}(D^S_{i2t}, D^{GT}) \\ 
        &+ \mathcal{F}_{KL}(D^S_{t2i}, D^{GT})
    \end{split}
\end{equation}
Moreover, $\mathcal{L}_{GT}$ is also called a hard target in the study~\citep{hinton2015distilling}. This can be viewed as a baseline for training dual-stream models without teacher distillation.

\textbf{Dual-stream distribution distillation.}
In this form, CLIP is often used as the dual-stream teacher, and the distribution of CLIP output is denoted as $D^{DS}_{i2t}, D^{DS}_{t2i}$. Then, a uniform loss function for dual-stream distillation is introduced as follows:
\begin{equation}
    \begin{split}
        \mathcal{L}_{TDD}: \mathcal{L}_{CLIP} = f_{DS}(D^S_{i2t},D^S_{t2i}) \\= \mathcal{F}_{KL}(D^S_{i2t}, D^{DS}_{i2t}) +\mathcal{F}_{KL}(D^S_{t2i},D^{DS}_{t2i}).
    \end{split}
\label{eq:dual_stream_dist_distill}
\end{equation}
This loss has been widely used in pioneering studies, including Leaner and Faster~\cite{ren2022leaner} which combin $\mathcal{L}_{GT}$ and $\mathcal{L}_{CLIP}$ as $\mathcal{L_{TDD}}$ and TinyCLIP~\cite{wu2023tinyclip} which solely uses $\mathcal{L}_{CLIP}$ as the distillation loss. We reimplement these works based on our model structures and datasets for fair comparisons.

%Then the loss function using only the dual-stream model, \textit{i.e.}, CLIP, can be denoted as follows:
%\begin{equation}
%    \begin{split}
%        \mathcal{L}_{TDD}: \mathcal{L}_{CLIP} &= f_{DS}(D^S_{i2t},D^S_{t2i})
%    \end{split}
%\end{equation}

\textbf{Single-stream distribution distillation.}
Similarly, ALBEF~\cite{li2021align} is also widely employed as the single-stream teacher. Then we construct the distribution of the ALBEF output by Eq.~(\ref{equ:single_stream_dist}). Due to the limitation of computing resources, we only calculate the top $k$ text similarities that are most similar to each image. Similarly, each text is treated in the same way. It should be noted that the information of top $k$ is provided by a dual-stream model, \textit{i.e.}, CLIP.
Then, the loss function using only the single-stream model, \textit{i.e.}, ALBEF, can be denoted as follows:
\begin{equation}
    \begin{split}
        &\mathcal{L}_{TDD}: \mathcal{L}_{ALBEF} =  \\   &\mathcal{F}_{KL}(\sigma(D^S_{i2t}[P_{i2t}]), \sigma(D^{SS}_{i2t}[P_{i2t}]))\\
    +&\mathcal{F}_{KL}(\sigma(D^S_{t2i}[P_{t2i}]), \sigma(D^{SS}_{t2i}[P_{t2i}])).\\
    \end{split}
\end{equation}
In LoopITR~\cite{lei2022loopitr} and Thinking Fast and Slow~\cite{miech2021thinking}, the authors employ $\mathcal{L}_{ALBEF} + \mathcal{L}_{GT}$ as $\mathcal{L_{TDD}}$ and we also implement them as comparison methods.
%So the output of the single-stream model creates the drawback that we only focus on the top k sample pairs and do not pay attention to the remaining negative samples.

\textbf{Dual-stream feature distillation.} In terms of target feature distillation, the dual-stream model outputs the image representation $I^{DS}$ and text presentations $ T^{DS}$ separately. Then, we can align the student's feature with the teacher's feature by constructing the following equation:
\begin{equation}
    \begin{split}
        & D^{FAI'}_{i2t} = \mathcal{F}_{D}(I^S, T^{DS}, (\tau^{DS}+\tau^S)/2) \\
        & D^{FAI'}_{t2i} = \mathcal{F}_{D}(T^{DS}, I^S, (\tau^{DS}+\tau^S)/2) \\
        & D^{FAT'}_{i2t} = \mathcal{F}_{D}(I^{DS}, T^S, (\tau^{DS}+\tau^S)/2) \\
        & D^{FAT'}_{t2i} = \mathcal{F}_{D}(T^S, I^{DS}, (\tau^{DS}+\tau^S)/2) \\
        &\mathcal{L}_{TFD}\!:\! \mathcal{L}_{CLIP\_FA} \!=\! f_{DS}(D^{FAI'}_{i2t},D^{FAI'}_{t2i})\\
        & \!+\!f_{DS}(D^{FAT'}_{i2t},D^{FAT'}_{t2i})
    \end{split}
\end{equation}
where $\mathcal{L}_{CLIP\_FA}$ indicates that we align the student features with the dual-stream teacher features, \textit{i.e.}, CLIP, and $f_{DS}$ is defined in Eq.~(\ref{eq:dual_stream_dist_distill}).

% \usepackage{multirow}
% \usepackage{graphicx}
\begin{table*}[tbh]
\centering
\setlength{\abovecaptionskip}{0.2em}
\setlength{\belowcaptionskip}{0.2em}
\caption{\small{Zero-shot comparisons and ablations on MSCOCO and Flickr30K testsets of the student model that uses the mobileViTv2 and TinyBERT as backbones. * indicates that we reimplement the comparable approaches based on our model structures and datasets. \# denotes the value of ALBEF are copied from the original paper.}}
\resizebox{\textwidth}{!}{%
\begin{tabular}{l|ll|llllllllllll}
\hline
\multirow{3}{*}{methods}                & \multicolumn{2}{c|}{Loss}                          & \multicolumn{6}{c}{\textbf{Flickr30K}}                                                                & \multicolumn{6}{c}{\textbf{MSCOCO}}                                                                    \\
                                        & \multicolumn{2}{c|}{$\mathcal{L}_{total} = \mathcal{L_{TDD}+L_{TFD}}$} & \multicolumn{3}{c}{\textbf{Image Retrieval}}           & \multicolumn{3}{c}{\textbf{Text Retrieval}}            & \multicolumn{3}{c}{\textbf{Image Retrieval}}           & \multicolumn{3}{c}{\textbf{Text Retrieval}}            \\
                                        & $\mathcal{L_{TDD}}$                  & $\mathcal{L_{TFD}}$             & R@1           & R@5           & R@10          & R@1           & R@5           & R@10          & R@1           & R@5           & R@10          & R@1           & R@5           & R@10          \\ \hline
CLIP                                 & $\mathcal{L}_{GT}$                   &                       & 85.2          & 97.5          & 99.1          & 64.9          & 87.2          & 92.1          & 56.3          & 79.4          & 86.7          & 36.7          & 61.4          & 71.5          \\
ALBEF \#                                & \multicolumn{2}{c|}{loss of ALBEF}                 & 94.1          & 99.5          & 99.7          & 82.8          & 96.3          & 98.1          & -             & -             & -             & -             & -             & -             \\ \hline
no teachers                             & $\mathcal{L}_{GT}$                   & -                     & 43.5          & 70.5          & 78.9          & 31.0          & 58.1          & 69.2          & 21.9          & 44.4          & 56.9          & 16.4          & 37.0          & 58.6          \\
\cite{lei2022loopitr}* & $\mathcal{L}_{ALBEF}+\mathcal{L}_{GT}$         & -                     & 54.3          & 81.1          & 88.2          & 40.5          & 70.2          & 79.3          & 30.5          & 56.5          & 67.2          & 21.5          & 45.6          & 57.5          \\
\cite{ren2022leaner}*  & $\mathcal{L}_{CLIP}+\mathcal{L}_{GT}$          & -                     & 56.2          & 80.2          & 87.9          & 40.5          & 67.8          & 77.2          & 29.7          & 55.9          & 67.5          & 20.7          & 43.4          & 55.0          \\
\cite{wu2023tinyclip}* & $\mathcal{L}_{CLIP}$                 & -                     & 60.1          & 82.4          & 89.2          & 41.3          & 69.2          & 78.5          & 31.9          & 58.1          & 67.9          & 21.1          & 44.3          & 56.0          \\ \hline
\multirow{8}{*}{Ablations of ours}      & $\mathcal{L}_{CLIP}$                 & -                     & 61.3          & 84.6          & 90.9          & 43.6          & 71.0          & 80.5          & 33.8          & 59.8          & 70.8          & 21.8          & 45.2          & 57.1          \\
                                        & $\mathcal{L}_{ALBEF}$                & -                     & 39.7          & 69.2          & 77.8          & 29.9          & 58.0          & 69.6          & 22.5          & 46.9          & 60.1          & 15.5          & 36.4          & 48.2          \\
                                        & $\mathcal{L}_{MT}$                   & -                     & 63.5          & 85.9          & 91.7          & 47.9          & 76.7          & 84.3          & 37.6          & 63.3          & 74.6          & 25.7          & 51.2          & 62.8          \\
                                        & -                          & $\mathcal{L}_{CLIP\_FA}$         & 60.5          & 83.8          & 89.5          & 41.9          & 69.9          & 79.3          & 32.6          & 57.7          & 68.7          & 21.1          & 43.6          & 55.2          \\
                                        & -                          & $\mathcal{L}_{MT\_FA}$           & 61.3          & 85.0          & 90.9          & 46.0          & 74.7          & 83.6          & 37.1          & 63.4          & 73.6          & 25.4          & 50.3          & 62.1          \\
                                        & $\mathcal{L}_{CLIP}$                 & $\mathcal{L}_{CLIP\_FA}$         & 64.3          & 87.0          & 92.2          & 46.2          & 74.3          & 82.7          & 35.7          & 62.4          & 72.6          & 23.5          & 46.9          & 58.7          \\
                                        & $\mathcal{L}_{CLIP}$                 & $\mathcal{L}_{MT\_FA}$           & 65.2          & 87.3          & 92.5          & 50.3          & 77.6          & 85.1          & 37.2          & 64.0          & 73.9          & 26.0          & 51.9          & 62.9          \\
                                        & $\mathcal{L}_{MT}$                   & $\mathcal{L}_{CLIP\_FA}$         & 66.0          & \textbf{88.1} & \textbf{92.7} & 51.1          & 78.3          & 85.7          & 37.7          & 64.4          & 74.6          & 26.6          & 52.1          & 63.3          \\ \hline
Ours                                    & $\mathcal{L}_{MT}$                   & $\mathcal{L}_{MT\_FA}$           & \textbf{66.6} & 87.4          & \textbf{92.7} & \textbf{52.1} & \textbf{78.9} & \textbf{86.6} & \textbf{38.6} & \textbf{65.0} & \textbf{75.2} & \textbf{27.3} & \textbf{52.7} & \textbf{64.0} \\ \hline
\end{tabular}%
}
\label{tab:result_on_tiny_bert}
\end{table*}

\begin{table*}[htb!]
\centering
\setlength{\abovecaptionskip}{0.2em}
\setlength{\belowcaptionskip}{0.2em}
\caption{\small{Finetuing ablations on MSCOCO and Flickr30K testsets of the student model that uses the MobileViTv2 and TinyBERT as backbones. \# denotes the value of ALBEF are copied from the original paper.}}
\resizebox{\textwidth}{!}{%
\begin{tabular}{@{}l|ll|llllllllllll@{}}
\toprule
\multirow{3}{*}{methods}           & \multicolumn{2}{c|}{Loss}                          & \multicolumn{6}{c}{\textbf{Flickr30K}}                                                                & \multicolumn{6}{c}{\textbf{MSCOCO}}                                                                    \\
                                   & \multicolumn{2}{c|}{$\mathcal{L}_{total} = \mathcal{L_{TDD}+L_{TFD}}$} & \multicolumn{3}{c}{\textbf{Image Retrieval}}           & \multicolumn{3}{c}{\textbf{Text Retrieval}}            & \multicolumn{3}{c}{\textbf{Image Retrieval}}           & \multicolumn{3}{c}{\textbf{Text Retrieval}}            \\
                                   & $\mathcal{L_{TDD}}$              & $\mathcal{L_{TFD}}$                 & R@1           & R@5           & R@10          & R@1           & R@5           & R@10          & R@1           & R@5           & R@10          & R@1           & R@5           & R@10          \\ \midrule
CLIP                               & $\mathcal{L}_{GT}$               & -                         & 95.3          & 99.7          & 100.0         & 84.0          & 97.0          & 98.7          & 74.2          & 92.3          & 96.0          & 57.3          & 81.8          & 88.7          \\
ALBEF \#                            & \multicolumn{2}{c|}{loss of ALBEF}                 & 95.9          & 99.8          & 100.0         & 85.6          & 97.5          & 98.9          & 77.6          & 94.3          & 97.2          & 60.7          & 84.3          & 90.5          \\ \midrule
\multirow{3}{*}{Ablations of ours} & -                      & $\mathcal{L}_{CLIP\_FA}$             & 75.5          & 92.7          & 96.9          & 59.1          & 85.3          & 91.4          & 48.8          & 75.4          & 84.3          & 36.5          & 65.8          & 76.8          \\
                                   & $\mathcal{L}_{CLIP}$             & -                         & 78.7          & 93.6          & 96.8          & 61.6          & 86.6          & 92.0          & 54.4          & 79.7          & 88.1          & 39.0          & 68.0          & 78.8          \\
                                   & $\mathcal{L}_{CLIP}$             & $\mathcal{L}_{CLIP\_FA}$             & 79.2          & 94.3          & 97.4          & 62.8          & 87.3          & 92.6          & 54.0          & 79.9          & 87.8          & 39.3          & 68.1          & 78.8          \\ \midrule
Ours                               & $\mathcal{L}_{MT}$               & $\mathcal{L}_{MT\_FA}$               & \textbf{80.2} & \textbf{95.9} & \textbf{97.8} & \textbf{64.1} & \textbf{88.4} & \textbf{93.4} & \textbf{55.0} & \textbf{80.4} & \textbf{88.2} & \textbf{40.2} & \textbf{69.2} & \textbf{79.5} \\ \bottomrule
\end{tabular}%
}
\label{tab:resulf_on_full_data}
\end{table*}

\textbf{Multi-teacher distillation.}
Our motivation for integrating the multi-teachers' output distributions is to gain a better distribution to distill the student model. Since single-stream models tend to perform better than dual-stream models, we argue that single-stream models can better distinguish difficult samples that cannot be discriminated against by dual-stream models. So, the final loss for measuring the distribution gap between student and multi-teacher is expressed as Eq.~(\ref{equ:multi_teacher_tdd}).
Furthermore, to align the student output features with the multi-teacher fused features, the loss function for feature alignment is expressed as Eq.~(\ref{equ:multi_teacher_tfd}).

\subsection{Zero-shot Experiments and Ablations}
For the student's image and text encoder, we utilize MobileViTv2~\cite{mehta2023separable} and TinyBERT~\cite{tinybert2020}, with 11.19 M and 14.71 M parameters, respectively. We also employ CLIP (ViT-L/14) and ALBEF as the teacher models, containing approximately 427.62M and 419.12M parameters, respectively. In order to evaluate the generalizability of the student model, we train it on both the CC3M and SBU datasets and subsequently assess its performance on the COCO and Flickr30k testsets. Moreover, the default value of $k$ is 11. All zero-shot results for comparisons with the aforementioned baselines and ablation studies are obtained using the checkpoint associated with the highest validation performance and presented in Table~\ref{tab:result_on_tiny_bert}.

As mentioned in Sec.~\ref{exp:baseline_component}, we propose a uniform loss paradigm for image-text retrieval distillation approaches. For fair comparisons, we reimplement several baseline methods based on the same model structures and datasets. Specifically, for TinyCLIP~\cite{wu2023tinyclip}, we adopt its \textit{affinity mimicking} loss (equivalent to $\mathcal{L}_{CLIP}$) and \textit{uniformly manual inheritance} for the TinyBERT text encoder. For Leaner and Faster~\cite{ren2022leaner}, we adopt its $\mathcal{L}_{CLIP}\!+\! \mathcal{L}_{GT}$ loss while eliminating the $\mathcal{L}_{HN}$ item since it's orthogonal to distillation approaches. For LoopITR~\cite{lei2022loopitr}, we employ its $\mathcal{L}_{ALBEF}\!+\!\mathcal{L}_{GT}$ as the loss function. In addition, we also conduct comprehensive ablation studies based on our loss components, as illustrated in Table~\ref{tab:result_on_tiny_bert}.

Several observations can be drawn from the statistics. 1) Compared to training without teachers ($\mathcal{L}_{GT}$), the CLIP target distribution distillation ($\mathcal{L}_{CLIP}$) can bring more effective information, but the result will not be further improved when combining them together ($\mathcal{L}_{CLIP}\!+\!\mathcal{L}_{GT}$). This indicates that the ground truth (usually very noisy) is not a good distribution when distilling the student model. 2) When we solely use the distribution of top $k$ output by ALBEF ($\mathcal{L}_{ALBEF}$), it does not work very well, revealing that we also need to take into account distributions of more negative samples. When we use both the distribution of ALBEF output and the ground truth ($\mathcal{L}_{ALBEF}\!+\!\mathcal{L}_{GT}$), the results are much better than using the ground truth alone, which shows that it is necessary to readjust the distribution of top $k$. 3) When you combine the distribution of multi-teachers ($\mathcal{L}_{MA}$), it is more effective than any single teacher.
4) Moreover, when distillation on both feature and output distribution of CLIP ($\mathcal{L}_{CLIP}\!+\!\mathcal{L}_{CLIP\_FA}$), it works better than distillation using only the similarity distribution ($\mathcal{L}_{CLIP}$), which demonstrates that aligning the student's features to the teacher's features improves student performance. 5) Furthermore, the best results are achieved when using a multi-teacher distribution and aligning the student features to the fused multi-teacher features ($\mathcal{L}_{MA}\!+\!\mathcal{L}_{MA\_FA}$). This is a good proof of the effectiveness of our multi-teacher cross-modal alignment distillation framework since both target distribution distillation and target feature distillation are important.
\begin{table*}[!ht]
\centering
\setlength{\abovecaptionskip}{0.2em}
\setlength{\belowcaptionskip}{0.2em}
\caption{\small{
Zero-shot performance on MSCOCO and Flickr30K testsets by employing MobileViTv3 and ALBERT as image and text encoder, respectively.}
} 
\scalebox{0.75}{
\begin{tabular}{cc|cccccccccccc}
\hline
\multicolumn{2}{c|}{\multirow{2}{*}{$\mathcal{L}_{total} = \mathcal{L_{TDD}} $+$ \mathcal{L_{TFD}}$}}& \multicolumn{6}{c}{\textbf{Flickr30K}}& \multicolumn{6}{c}{ \textbf{MSCOCO}} \cr && \multicolumn{3}{c}{\textbf{Image Retrieval}}
& \multicolumn{3}{c}{\textbf{Text Retrieval}} & \multicolumn{3}{c}{\textbf{Image Retrieval}} & \multicolumn{3}{c}{\textbf{Text Retrieval}} \\

$\mathcal{L_{TDD}}$& $\mathcal{L_{TFD}}$&R@1 &R@5 &R@10  &R@1 &R@5 &R@10 &R@1 &R@5 &R@10 &R@1 &R@5 &R@10 \\
\hline
% \makecell[r]{ mobile\_clip +$\mathcal{L}_{Student}$}
% &43.2&72.6&82.9&30.5&56.9&68.3&23.4&47.3&59.4&16.9&38.0&49.9\\
% \hline

% \makecell[l]{mobile\_clip +$\mathcal{L}_{Student}$ \\ +$\mathcal{L}_{CLIP}$}
%   &57.0&82.6&89.4&40.2&66.9&76.8&32.5&58.0&69.4&21.1&44.8&56.6\\
% \hline
% \makecell[l]{mobile\_clip  +$\mathcal{L}_{CLIP}$ {\color{red}}}
%   &62.2&85.0&91.5&44.2&72.8&82.0&34.9&60.3&71.2&23.2&47.1&58.9\\
% \hline
\makecell[l]{ $\mathcal{L}_{CLIP}$ 
 }&$\mathcal{L}_{CLIP\_FA}$&62.0&86.2&91.8&45.7&73.9&82.4&35.7&62.0&72.7&23.4&48.1&59.9\\

\makecell[l]{ $\mathcal{L}_{MT}$}&$\mathcal{L}_{MT\_FA}$&\textbf{64.8}&\textbf{88.0}&\textbf{93.9}&\textbf{49.6}&\textbf{77.5}&\textbf{85.8}&\textbf{35.9}&\textbf{63.3}&\textbf{74.5}&\textbf{26.0}&\textbf{51.6}&\textbf{63.4}\\

\hline
\end{tabular}}
\label{tab:result_on_albert}
\end{table*}

\begin{table*}[h!]
\centering
\setlength{\abovecaptionskip}{0.2em}
\setlength{\belowcaptionskip}{0.2em}
\caption{\small{The student model's performance is assessed with various hyper-parameters $k$ through zero-shot evaluations on MSCOCO and Flickr30K testsets by using mobileViTv2 and TinyBERT as image and text encoder, respectively.}
} 
\scalebox{0.8}{
\begin{tabular}{l|cccccccccccc}
\hline
\multirow{3}{*}{$\mathcal{L_{TDD}}$: $\mathcal{L}_{MA}$
}& \multicolumn{6}{c}{\textbf{Flickr30K}}& \multicolumn{6}{c}{ \textbf{MSCOCO}} \cr & \multicolumn{3}{c}{\textbf{Image Retrieval}}
& \multicolumn{3}{c}{\textbf{Text Retrieval}} & \multicolumn{3}{c}{\textbf{Image Retrieval}} & \multicolumn{3}{c}{\textbf{Text Retrieval}} \\
 $\mathcal{L_{TFD}}$: $\mathcal{L}_{MA\_FA}$&R@1 &R@5 &R@10  &R@1 &R@5 &R@10 &R@1 &R@5 &R@10 &R@1 &R@5 &R@10 \\
\hline
% \makecell[r]{ mobile\_clip +$\mathcal{L}_{Student}$}
% &43.2&72.6&82.9&30.5&56.9&68.3&23.4&47.3&59.4&16.9&38.0&49.9\\
% \hline

% \makecell[l]{mobile\_clip +$\mathcal{L}_{Student}$ \\ +$\mathcal{L}_{CLIP}$}
%   &57.0&82.6&89.4&40.2&66.9&76.8&32.5&58.0&69.4&21.1&44.8&56.6\\
% \hline
% \makecell[l]{mobile\_clip  +$\mathcal{L}_{CLIP}$ {\color{red}}}
%   &62.2&85.0&91.5&44.2&72.8&82.0&34.9&60.3&71.2&23.2&47.1&58.9\\
% \hline
\makecell[c]{$k=5$}
  &64.3&\textbf{87.5}&92.6&50.8&78.1&85.9&\textbf{38.8}&65.0&74.8&26.6&52.0&63.5\\

\makecell[c]{$k=11$}
  &\textbf{66.6}&{87.4}&92.7&\textbf{52.1}&\textbf{78.9}&\textbf{86.6}&{38.6}&\textbf{65.0}&\textbf{75.2}&\textbf{27.3}&\textbf{52.7}&\textbf{64.0}\\

\makecell[c]{$k=17$}
  &63.3&86.9&\textbf{93.0}&50.6&78.6&86.0&37.5&64.6&75.1&26.6&52.1&63.5\\
\hline

\end{tabular}}
\label{tab:result_on_numk}
\end{table*}

\begin{figure*}[!ht]
	\centering
 \includegraphics[width=0.95\textwidth]{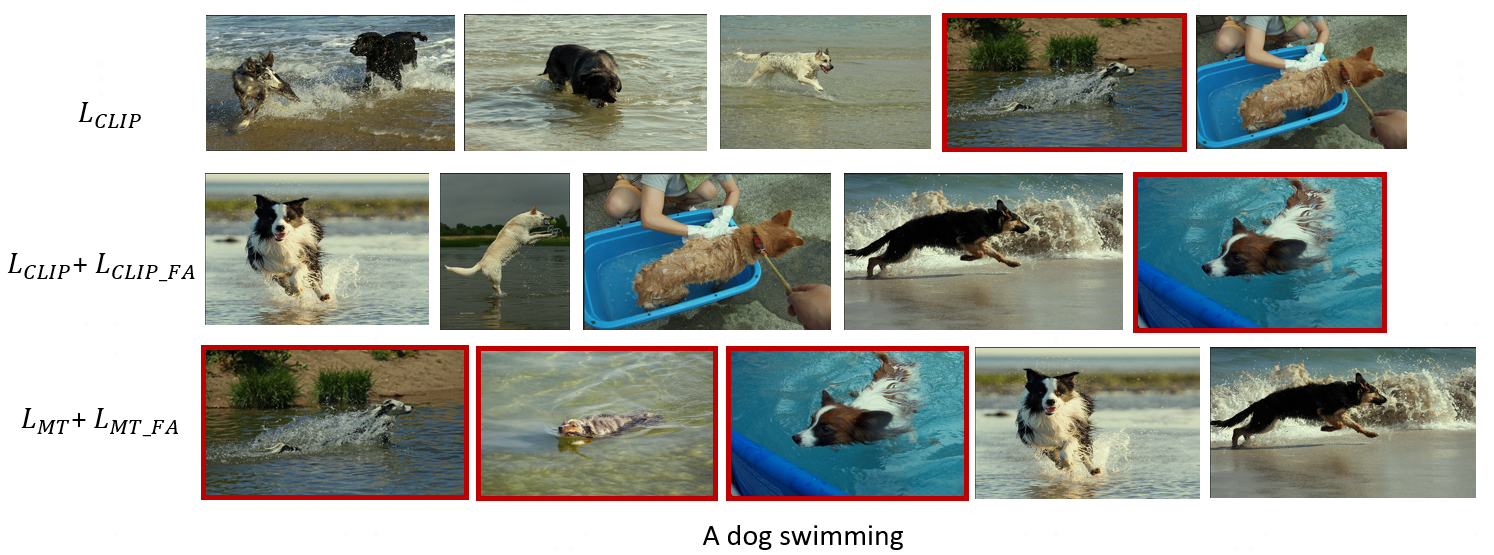}
 \includegraphics[width=0.95\textwidth]{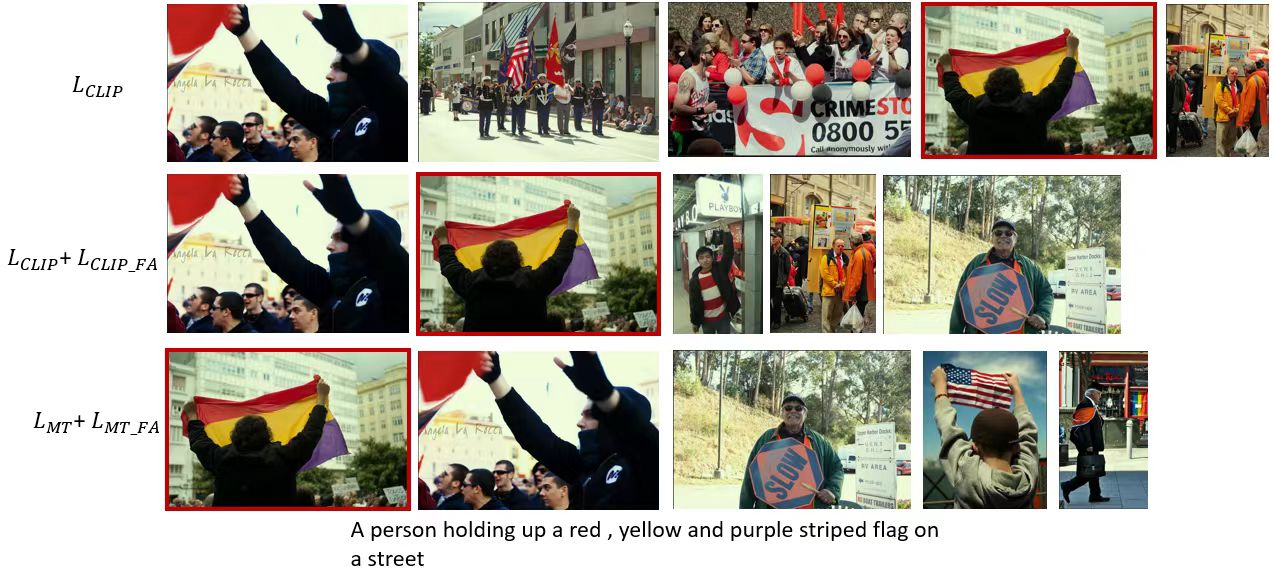}
 \includegraphics[width=0.95\textwidth]{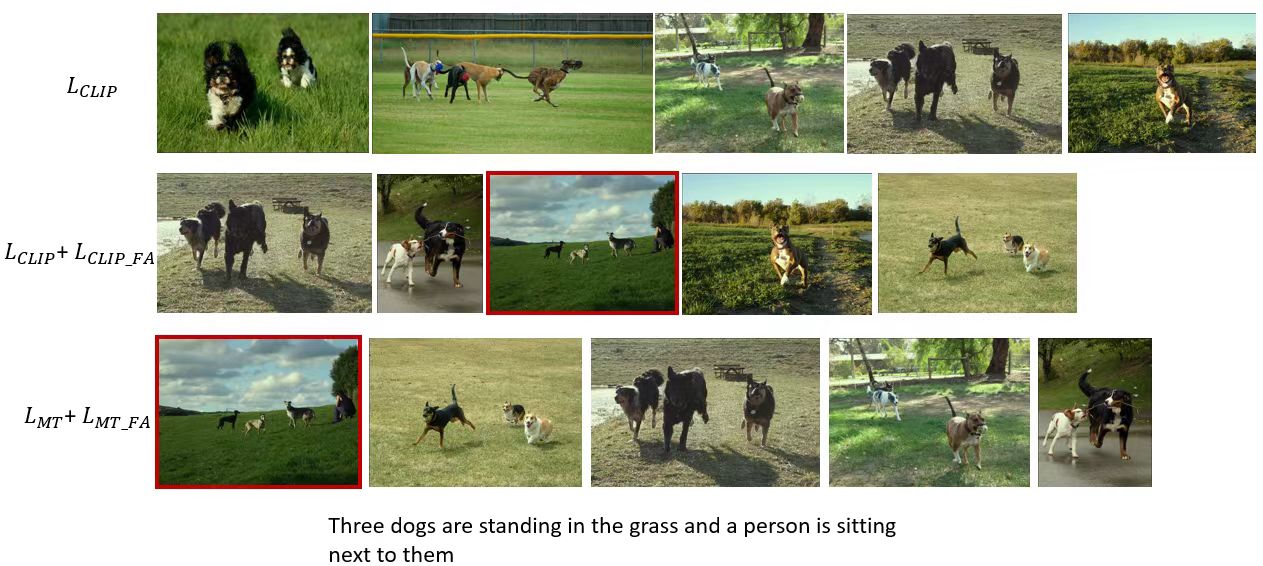}
	\caption{Retrieval results obtained using different distillation methods}
	\label{fig:visual_results}
\end{figure*}

To further illustrate the impact of different distillation methods, we select several text-to-image retrieval results (Flickr30k testset) for three different methods, i.e. $\mathcal{L}_{CLIP}$, $\mathcal{L}_{CLIP}\!+\!\mathcal{L}_{CLIP\_FA}$,$\mathcal{L}_{MA}\!+\!\mathcal{L}_{MA\_FA}$ as described in Table~\ref{tab:result_on_tiny_bert}, and those visualization results are shown in Figure~\ref{fig:visual_results}. We can observe that our MCAD ($\mathcal{L}_{MA}\!+\!\mathcal{L}_{MA\_FA}$) achieves more accurate matching results for fine-grained attribute words, such as action ("swimming"), color ("red, yellow, and purple" ) and number ("three"). Since CLIP is not good at discriminating between such subtle differences because of its shallow image-text interaction module, which has been mentioned in many pinoneer work~\cite{Doveh_2023_CVPR}, we believe that such improvement is distilled from the single-stream teacher (ALBEF).

\begin{table*}[h!]
\centering
\setlength{\abovecaptionskip}{0.2em}
\setlength{\belowcaptionskip}{0.2em}
\caption{\small{Scan speed, retrieval speed and running memory of different models based on 100,000 candidate images. * indicates that ALBEF selects the top 128 candidates for the fusion module calculation.}}
\resizebox{\textwidth}{!}{%
\begin{tabular}{@{}l|llll|llll@{}}
\toprule
Model                                      & image encoder                    & text encoder              & fusion module                & param. \#               & scan time & retrie. time & running mem. & platform         \\ \midrule
\multicolumn{1}{l|}{CLIP}                  & VIT-L/14                         & 12-layer transformer      & dot product                  & 427.62M                 & 11.0ms              & 32.5ms         & $\sim$2GB      & V100 GPU         \\
\multicolumn{1}{l|}{ALBEF}                 & VIT-B/16                         & 6-layer transformer       & 6-layer transformer          & 419.12M                 & 7.6ms               & 1945ms*        & $\sim$3GB      & V100 GPU         \\ \midrule
\multicolumn{1}{l|}{\multirow{3}{*}{ours}} & \multirow{3}{*}{mobileVitV2-1.5} & \multirow{3}{*}{TinyBERT} & \multirow{3}{*}{dot product} & \multirow{3}{*}{25.9 M} & 3.8ms               & 14.1ms         & $\sim$150MB    & V100 GPU         \\
\multicolumn{1}{l|}{}                      &                                  &                           &                              &                         & 24.5ms              & 8.5ms          & 93MB           & Snapdragon 8 Gen3 \\
\multicolumn{1}{l|}{}                      &                                  &                           &                              &                         & 24.8ms              & 7.5ms          & 107MB          & Dimensity 9300   \\ \bottomrule
\end{tabular}}
\label{tab:Performance_on_mobile_phones}
\end{table*}

\subsection{Finetuning Experiments}
To further verify the finetuning performance of our approach, we first finetune the teacher models and then perform different distillation strategies on the MSCOCO and Flickr30K training datasets. All results are shown in Table~\ref{tab:resulf_on_full_data}, which maintains the same conclusion as before. We achieve the best results when combining multi-teacher distribution distillation ($\mathcal{L}_{MT}$) and feature distillation ($\mathcal{L}_{MT\_FA}$), surpassing dual-stream distribution distillation ($\mathcal{L}_{CLIP}$), feature distillation ($\mathcal{L}_{CLIP\_FA}$) and combining them together ($\mathcal{L}_{CLIP}+\mathcal{L}_{CLIP\_FA}$).

\subsection{Backbone and Hyper-parameter Selection}
% \begin{table*}[!ht]
% \centering
% \scalebox{0.8}{
% \begin{tabular}{l|cccccccccccc}
% \hline
% \multirow{3}{*}{\textbf{model}}& \multicolumn{6}{c}{\textbf{Flickr30K}}& \multicolumn{6}{c}{ \textbf{MSCOCO}} \cr & \multicolumn{3}{c}{\textbf{Image Retrieval}}
% & \multicolumn{3}{c}{\textbf{Text Retrieval}} & \multicolumn{3}{c}{\textbf{Image Retrieval}} & \multicolumn{3}{c}{\textbf{Text Retrieval}} \\
%  ~&R@1 &R@5 &R@10  &R@1 &R@5 &R@10 &R@1 &R@5 &R@10 &R@1 &R@5 &R@10 \\
% \hline
% CLIP (origin)&85.2&97.5&99.1&64.9&87.2&92.1&56.3&79.4&86.7&36.7&61.4&71.5\\
% \hline
% CLIP (after fine-tuning)
%   &95.3&99.7&100.0&84.0&97.0&98.7&74.2&92.3&96.0&57.3&81.8&88.7\\
% \hline

% ALBEF (origin)
%   &94.1&99.5&99.7&82.8&96.3&98.1&-&-&-&-&-&-\\
% \hline

% ALBEF (after fine-tuning)
%   &95.9&99.8&100.0&85.6&97.5&98.9&77.6&94.3&97.2&60.7&84.3&90.5\\
% \hline

% \end{tabular}}
% \caption{
% The Performance of Teacher Models Before and After Fine-tuning. The experimental results of ALBEF were obtained from \citep{li2021align}  }

% \label{tab:teachers_performance}
% \end{table*}

Since our method is a network-agnostic framework, we replace the image encoder and text encoder with MobileViTv3~\citep{wadekar2022mobilevitv3} and ALBERT~\citep{Lan2020ALBERT}, with 5.5M and 12.2M parameters, respectively, to validate its generality. All zero-shot results are shown in Table~\ref{tab:result_on_albert}. The statistics show that our approach still outperforms $\mathcal{L}_{CLIP}+\mathcal{L}_{CLIP\_FA}$,  revealing that our proposed method is general to different dual-stream models.

Further, we conduct several experiments on the selection of $k$, with results shown in Table~\ref{tab:result_on_numk}, which illustrates the impact of the hyper-parameter $k$ on the distillation effect. Specifically, a lower R@1 score for the Flickr30k data is observed when $k$ is set to 5 due to the diminished information received from ALBEF. Conversely, when $k$ is increased to 17, the distribution of information from ALBEF becomes smoother, impeding the student model's ability to learn more accurate information. Notably, this aforesaid effect is most pronounced in the R@1 scores. Therefore, it is essential to select an appropriate value of $k$ to enhance the performance of the student model. Finally, we choose an optimal $k=11$ for all experiments.

\subsection{Mobile-device Application}
Table~\ref{tab:Performance_on_mobile_phones} tests the performance of the lightweight model deployed on Snapdragon 8 Gen3 and MTK Dimensity 9300 chips, which uses TinyBERT as the text encoder and mobileViTv2 as the image encoder that builds an offline index using 100,000 candidate images. We successfully achieve $\sim$24.6ms/image scan speed, $\sim$8.0ms/query real-time retrieval speed, and $\sim$100MB running memory. Thanks to the deep optimization on the chip side, the retrieval speed even surpasses that on the V100 GPU, greatly advancing the mobile-device application of VLP models.

\section{Conclusion and Liminations}
In this study, we propose a multi-teacher cross-modal alignment distillation (MCAD) framework which helps better integrate heterogeneous teachers. The proposed MCAD involves the integration of the teachers' output features and similarity distributions. Moreover, MCAD uses the integrated distributions to distill the student model and align the student's features to the fused teachers' features. Our proposed MCAD is demonstrated to be a model-agnostic general framework, capable of achieving superior performance on both zero-shot and fine-tuning settings and a lightweight model has been successfully deployed on mobile devices, achieving real-time retrieval speed. 

In this research, due to computational resource constraints, we only conduct a few experiments and simply determine the hyper-parameter $k=11$ for all experiments. But it may be dynamic for different datesets and networks. Another limitation is that using MLP as the projection layers in Eq.~(\ref{eq:projection_layer}) my not be optimal and more intricate designs need to be investigated in the future.

% \subsection{Footnotes}

% Footnotes are inserted with the \verb|\footnote| command.\footnote{This is a footnote.}

% \nocite{Ando2005,borschinger-johnson-2011-particle,andrew2007scalable,rasooli-tetrault-2015,goodman-etal-2016-noise,harper-2014-learning}

% \subsection{Appendices}

% Use \verb|\appendix| before any appendix section to switch the section numbering over to letters. See Appendix~\ref{sec:appendix} for an example.

% \commentYb{\begin{itemize}

%     \item The trade off between the ground truth and the soft label.
%     \item The selection of hyperparameter $k$.
%     \item How to combine the distributions obtained by different teachers
% \end{itemize}}
% \section*{Acknowledgements}

\newpage
\bibliography{anthology,custom}

\newpage
\appendix

\section{Dataset Statistics}
\label{app:dataset_statistics}
Some of the images are no longer accessible on the internet, and the CC dataset we collect for training is not quite complete. 
Table \ref{tab:dataset} shows the statistics of the datasets.  $\intercal$ in Table~\ref{tab:dataset} denotes the data we only used for the fine-tuning stage. 

\begin{table}
\centering
\caption{\small{
statistic of the dataset. $\intercal$ represents the datasets only used in the fine-tuning stage.}}
\begin{tabular}{c|cccc}
% \hline
\toprule[1pt]
\textbf{Datasets} &  CC & SUB & $\text{COCO}$ & $\text{Flickr}$\\
% \hline
\midrule[0.5pt]
train  & 1.90M & 0.85 M & $\text{0.56M}^\intercal$ & $\text{0.15M}^\intercal$\\
val  & - & - & 25k& 5k \\
test & - & - & 25k & 5k\\
% \hline
\bottomrule[1pt]
\end{tabular}
\label{tab:dataset}
\end{table}

\section{Train Details}
\label{app:training_details}

% For all experiments, we use AdamW~\citep{loshchilov2017decoupled} with $\beta_1=0.9, \beta_2 =0.999$ for optimization. The default learning rate is 1e-3 except when testing Albert+ MobileVitv3 as backbone, which is set to 1e-4. And warm-up with cosine decay are used. To speed up the training, we use apex framework. For teacher models in all experiments, We do not use any data augmentation techniques. But for student, we use only ``RandomResizedCrop" operation.   To make the model training more adequate, 100 epochs are trained for each experiment.
% Throughout this paper, the value of the hyperparameter k is 11 unless otherwise specified.

In this study, the AdamW~\citep{loshchilov2017decoupled} optimization technique with $lr=1e-3, \beta_1=0.9, \beta_2 =0.999$ is employed for all experiments, except for the test on ALBERT+ MobileVitv3 backbone, where the default learning rate is adjusted to 1e-4. To facilitate the training process and enhance the performance, warm-up with cosine decay is applied, while the apex framework is utilized to accelerate the training. Notably, no data augmentation methods are utilized in the teacher models, while the student model only employs "RandomResizedCrop". Moreover, to ensure sufficient training, each experiment is trained for 100 epochs. It is important to mention that the value of the hyperparameter k is set to 11, unless otherwise specified in this paper.

In terms of teacher models, we adopt CLIP ViT-L/14 as the dual-stream teacher\footnote{\url{https://huggingface.co/openai/clip-vit-large-patch14}} and ALBEF\footnote{\url{https://github.com/salesforce/ALBEF}} as the single-stream teacher. For the projection layers $\{g_1, g_2, g_3, g_4\}$, we simply employ two-layer MLPs.

\section{Loss Explanation}
\label{app:loss_explanation}
We choose a special form of normalization term in Eq.~(\ref{eq:L1norm}). We can view it as an L1 normalization. Here we want to explain why we choose such normalization formulation instead of commonly used softmax. Given that single-stream models similar to ALBEF typically output a score for an image-text pair, to ensure that the scores of different sample pairs maintain their relative magnitude after normalization, we employed this specific normalization approach. Take an example for clearer clarification. Assume that the top 3 output scores are {0.8, 0.4, 0.2} (the probability of two-classification after ALBEF must be between 0 and 1), after the L1 normalization, the outputs are $\{\frac{0.8}{0.8+0.4+0.2}\}$ = $\{0.571, 0.286, 0.143\}$. The relative ratio is still $4:2:1$. But if we choose softmax normalization, the output becomes $\{0.451, 0.302, 0.247\}$, which is much smoother and lacking in differentiation. Actually, we have indeed tried to apply softmax normalization during our experiments, but we found that simply using a softmax would cause ALBEF's score distribution to become smoother and result in inferior performance, while incorporating a temperature-scaled softmax function would introduce additional hyper-parameters. So we finally chose the L1 normalization method.

\end{document}